\documentclass{article}

\usepackage[nonatbib,final]{neurips_2018}

\usepackage[utf8]{inputenc} % allow utf-8 input
\usepackage[T1]{fontenc}    % use 8-bit T1 fonts
\usepackage[colorlinks=true,linkcolor=blue,citecolor=blue]{hyperref}       % hyperlinks
\usepackage{url}            % simple URL typesetting
\usepackage{booktabs}       % professional-quality tables
\usepackage{amsfonts}       % blackboard math symbols
\usepackage{nicefrac}       % compact symbols for 1/2, etc.
\usepackage{microtype}      % microtypography
\usepackage{graphicx}
\usepackage{helvet}
\usepackage{courier}
\usepackage{amsmath}
\usepackage{amssymb}
\usepackage{tabularx}
\usepackage{varwidth}
\usepackage{xcolor}
\usepackage{graphicx}
\usepackage{bbm}
\usepackage{bm}
\usepackage{colortbl}
\usepackage[clock]{ifsym}
\usepackage{enumitem}
\usepackage{multicol} % Required for multiple columns
\usepackage[utf8]{inputenc}
\usepackage{booktabs}  
\usepackage{algorithm}
\usepackage{algorithmic}
\usepackage{wrapfig}
\usepackage{amsthm}
\usepackage{multirow}
\usepackage{subfig}
\usepackage{setspace}
\usepackage{pifont}

\newcolumntype{C}[1]{>{\centering\arraybackslash}m{#1}}
\newcolumntype{R}[1]{>{\raggedright\arraybackslash}m{#1}}

\newcommand{\indep}{\raisebox{0.08em}{\rotatebox[origin=c]{90}{$\models$}}}
\newcommand{\notindep}{\not\!\perp\!\!\!\perp}
\newcommand*{\QED}{\hfill\ensuremath{\blacksquare}}

\usepackage[font=small,skip=10pt]{caption}
\usepackage{setspace}
\DeclareCaptionFormat{myformat}{#1#2#3\hrulefill}

\title{Accounting for Unobserved Confounding in Domain Generalization}

\author{
  Alexis Bellot$^{1,2}$\hspace{0.5cm} Mihaela van der Schaar$^{1,2,3}$\\
  $^{1}$University of Cambridge, $^{2}$The Alan Turing Institute, $^{3}$University of California Los Angeles\\
  \texttt{[abellot,mschaar]@turing.ac.uk} \\
}

\begin{document}

\maketitle

\begin{abstract}
This paper investigates the problem of learning robust, generalizable prediction models from a combination of multiple datasets and qualitative assumptions about the underlying data-generating model. Part of the challenge of learning robust models lies in the influence of unobserved confounders that void many of the invariances and principles of minimum error presently used for this problem. Our approach is to define a different invariance property of causal solutions in the presence of unobserved confounders which, through a relaxation of this invariance, can be connected with an explicit distributionally robust optimization problem over a set of affine combination of data distributions. Concretely, our objective takes the form of a standard loss, plus a regularization term that encourages partial equality of error derivatives with respect to model parameters. We demonstrate the empirical performance of our approach on healthcare data from different modalities, including image, speech and tabular data.
\end{abstract}

\section{Introduction}
Prediction algorithms use data, necessarily sampled under specific conditions, to learn correlations that extrapolate to new or related data. If successful, the performance gap between these two environments is small, and we say that algorithms \textit{generalize} beyond their training data. Doing so is difficult as the set of potential distributional changes at test time is mostly unknown and may be large and varied. This problem is broadly known
as dataset shift and requires a formalization of how dataset shift arises, and how that shift impacts the conditional distribution of our target $Y$ given features $\mathbf X$.

One common way to formalize learning with dataset shift is, instead of optimizing for correlations in a single data distribution, to do so simultaneously for multiple different distributions in an uncertainty set $\mathcal P$,
\begin{align}
\label{robust_pop}
    \underset{f}{\text{minimize }} \underset{P \in \mathcal P}{\sup}\hspace{0.1cm} \mathbb E_{(\mathbf x,y)\sim P} [ \mathcal L(f(\mathbf x),y)],
\end{align}
for some measure of error $\mathcal L$ of the function $f$ that relates input and output examples $(\mathbf x,y)\sim P$. Different sets $\mathcal P$ formalize the expected shifts and leads to estimators with different robustness properties and different performance for a given data distribution. Robust solutions to problem (\ref{robust_pop}) are said to generalize if potential shifted, test distributions are contained in $\mathcal P$, but also larger sets $\mathcal P$ result in conservative solutions (i.e. with sub-optimal performance) on data sampled from distributions away from worst-case scenarios. For instance, causal solutions have been shown to optimize for (\ref{robust_pop}) with $\mathcal P$ defined as any distribution arising from interventions in an underlying structural causal model on observed covariates $\mathbf x$ leading to shifts in their distribution $P_{\mathbf x}$ (see e.g. sections 3.2 and 3.3 in \cite{meinshausen2018causality} and \cite{peters2016causal}). The invariance to changes in covariate distributions of causal solutions is powerful for generalization and has been used to justify invariant risk minimization algorithms such as proposed in \cite{arjovsky2019invariant} and extensions \cite{ahuja2020invariant,krueger2020out,gimenez2020identifying, lu2021nonlinear, jin2020enforcing, magliacane2017domain,koyama2020out,parascandolo2020learning}. 

Invariant solutions across environments, their optimality for generalization, and their causal interpretation is valid however only under the premise that all covariates and drivers of the outcome subject to change at test time are observed. Often shifts occur elsewhere, for example in the distribution of unobserved confounders, in which case also conditional distributions $P_{y|\mathbf x}$ may shift, or on the target $Y$ itself. In the presence of unobserved confounders, the goals of achieving robustness and learning a causal model can be \textit{different}. There is, in general, an inherent \textit{trade-off} in generalization performance: in the presence of unobserved confounders, causal and correlation-based solutions are both optimal in different regimes, depending on the shift in the underlying data generating mechanism from which new data is generated. For additional context, we consider next a simple example that we re-use throughout this paper to highlight different aspects of this problem.

\begin{figure*}[t]
%\captionsetup{format=myformat}
\centering
\includegraphics[width=1\textwidth]{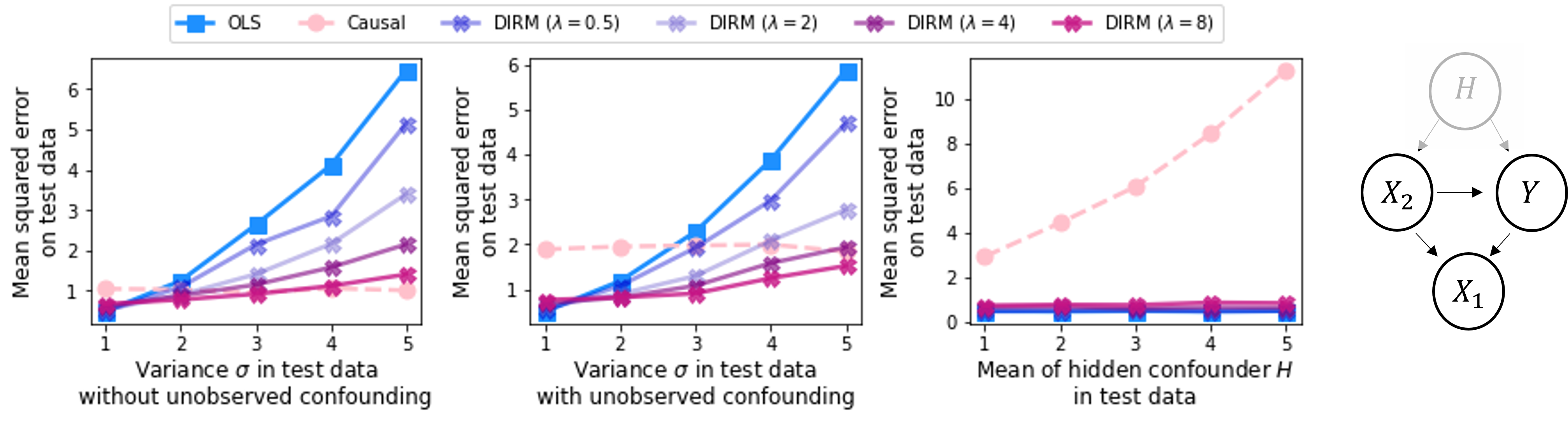}
\caption[.]{Each panel plots testing performance under different shifts. The proposed approach, Derivative Invariant Risk Minimization (DIRM, described in section \ref{sec_gen}), is a relaxation of the causal solution that interpolates as a function of a hyperparameter $\lambda$ (defined eq. (\ref{robust_objective})) between the causal solution and Ordinary Least Squares (OLS).}
\label{Fig1}
\end{figure*}

\subsection{Motivating example}
Assume access to observations of variables $(X_1,X_2,Y)$ in two training datasets, each dataset sampled with different variances ($\sigma^2=1$ and $\sigma^2 = 2$) from the following structural model,
\begin{align*}
    X_2 := -\eta H + (1-\eta)E_{X_2}, \quad Y := X_2 + \eta H + (1-\eta)E_{Y},\quad X_1 := Y + X_2 + E_{X_1}.
\end{align*}
$E_{X_1}, E_{X_2}\sim\mathcal N(0,\sigma^2)$, $E_Y\sim\mathcal N(0,1)$ are independent error terms and $H\sim\mathcal N(0,1)$ is an unobserved confounder whose influence is determined by $\eta\in\{0,1\}$\footnote{The motivation for including $\eta$ is to introduce unobserved confounding without increasing the variance of the observed variables.}. We consider two data generating scenarios. 
\begin{enumerate}[leftmargin=*, itemsep=0pt, topsep=0pt]
    \item \textbf{Leftmost panel} of Figure \ref{Fig1} all data is generated \textit{without} unobserved confounders, $\eta:=0$.
    \item \textbf{Middle and rightmost panel} of Figure \ref{Fig1} all data is generated \textit{with} unobserved confounders, $\eta = 1$.
\end{enumerate}
Each panel of Figure \ref{Fig1} shows performance on new data obtained after manipulating the underlying data generating system; the magnitude and type of intervention appears in the horizontal axis. We consider the minimum average error solution Ordinary Least Squares (OLS), the causal solution i.e. the linear model with coefficients $(0,1)$ for $(X_1,X_2)$, and Derivative Invariant Risk Minimization (DIRM, the proposed approach described in section \ref{sec_gen}) in different instantiations as a function of a hyperparameter $\lambda$). Two observations motivate this paper.

\begin{enumerate}[leftmargin=*,itemsep=0pt]
    \item Causal solutions optimize for worst case error under interventions on observed variables of arbitrary magnitude. By construction such solutions are typically not optimal under moderate interventions, especially in the presence of unobserved confounders, as shown in the middle panel. Moreover, by abstracting from modelling any dependency due to confounders performance rapidly deteriorates with the amount of variation in the outcome due to unobserved confounders, as shown in the rightmost panel.
    \item Minimum average error solutions by contrast optimize for any dependency (even if confounded) in the available data, whose performance as a result deteriorates on data subject to interventions on observed variables but better adjusts to interventions on unobserved confounders.  Minimum average error and causal solutions can be interpreted as two extremes of a distributionally robust optimization problem (\ref{robust_pop}), with a range of intermediate solutions that DIRM seeks to exploit and that in practice may have a more desirable performance profile. 
\end{enumerate}

\subsection{Our Contributions} 
This work investigates generalization performance in the presence of unobserved confounding with data from multiple environments. Section \ref{sec_2} emphasizes a qualitative difference in the statistical invariances that can be expected in the presence of unobserved confounders. We show that the trade-off in the performance of different objectives in the presence of unobserved confounders and the proposed invariance principles suggest a new objective, Derivative Invariant Risk Minimization (briefly illustrated in Figure \ref{Fig1} and described in section \ref{sec_gen}), that defines a range of intermediate solutions between the causal and minimum error extremes. 

Section \ref{sec_gen} shows that solutions to the Derivative Invariant Risk Minimization problem are robust in a well-defined sense, as upperbounding a robust minimization problem (\ref{robust_pop}) that defines $\mathcal P$ as an \textit{affine} combination of training data distributions. When the set of affine combinations $\mathcal P$ is interpreted as a set of distributions arising from shifts in an underlying structural causal model actually defines robustness in a broader sense than causal solutions, including robustness to interventions in unobserved and target variables with the caveat however that the geometry of training environments must present the shift of interest (see section \ref{sec_rob_inter}). Finally, we conclude this paper with a discussion of related work and with performance comparisons on medical data and other benchmarks for domain generalization.

\subsection{Problem formulation}
We introduce in this section some basic notations and definitions that will be used throughout the paper. We use capital letters to denote variables ($X$), small letters for their values ($x$), bold letters for sets of variables ($\mathbf X$) and their values ($\mathbf x$), and caligraphic letters for their domains ($\mathbf x\in\mathcal X$). We reserve $\mathcal P$ to denote collections of (indexed) probability distribution and $\mathcal E$ to denote collections of indices. For convenience, we denote by $P(\mathbf x)$ probabilities $P(\mathbf X = \mathbf x)$. All proofs are given in the Appendix.

The basic semantical framework of our analysis rests on structural causal models (SCMs) (Definition 7.1.1 \cite{pearl2009causality}). An SCM $M$ is a tuple $M = \langle \mathbf V , \mathbf U,\mathbf F, P \rangle$ where $\mathbf V$ is a set of endogenous variables and $\mathbf U$ is a set of exogenous variables. $\mathbf F$ is a set of functions where each $f_V \in \mathbf F$ decides values of an endogenous variable $V \in \mathbf V$ taking as argument a combination of other variables in the system. That is, $v := f_V (pa_V , u_V)$, with $Pa_V \in \mathbf V , U_V\in \mathbf U$. Exogenous variables $U\in\mathbf U$ are mutually independent\footnote{Unobserved confounding will arise from the fact that not all endogeneous variables $\mathbf V$ are observed in data, only a subset $\mathbf X \subset \mathbf V$ is observed.}, values of which are drawn from the exogenous distribution $P(\mathbf u)$. Naturally, $M$ induces a joint distribution $P(\mathbf v)$ over endogenous variables $\mathbf V$.

\textbf{Definition 1} (Additive invertible SCMs). \textit{An SCM $M = \langle \mathbf V , \mathbf U,\mathbf F, P \rangle$ is said to be additive and invertible if $\mathbf F$ is restricted to the set of additive, invertible functions, i.e. for any two inputs $v_i,v_j$ and $f\in\mathbf F$, $f(v_i + v_j) = f(v_i) + f(v_j)$ and $f(v_i) = v_j$ if and only if there exists $f^{-1}$ such that $f^{-1}(v_j) = v_i$.}

This restriction will allow us to define invariances between environments in the presence of unobserved confounders but will be relaxed for making arguments on distributional robustness in section \ref{sec_gen}.

\textbf{Definition 2} (Environments). \textit{An environment $i$ is defined as a distribution $P_i(\mathbf x)$ over a subset $\mathbf X \subset \mathbf V$ generated by an additive, invertible SCM $M = \langle \mathbf V , \mathbf U,\mathbf F, P_i \rangle$. Different environments differ in their distribution over exogeneous variables $P_i(\mathbf u_{\mathbf x})$ while the set of observed endogeneous variables $\mathbf X \subset \mathbf V$ and causal mechanisms $\mathbf F$ are invariant.}

We take the perspective that all distributions that may be observed over a system of variables $\mathbf X\subset \mathbf V$ arise from a single underlying additive, invertible SCM with potentially differing distribution over exogeneous variables which propagates across the system shifting the joint distribution of all variables. Representative real-world examples include data collected subject to changing data collection conditions, such as due to different measurement devices, or different data sources.

The problem we consider is to learn a prediction function $f:\mathcal X \rightarrow \mathcal Y$ that takes observations of variables $\mathbf X$ to approximate a target variable $Y$ from data in multiple environments, designed to generalize in a well-defined sense. Learned $f$ should not necessarily agree with $f_Y\in\mathbf F$ as causal solutions are not necessarily optimal in the presence of unobserved confounding.

\section{Biases and invariances with unobserved confounding}
\label{sec_2}

It will be useful to write $f_Y = g_Y \circ \phi_Y$ as a composition of a representation function $\phi_Y:\mathcal X \rightarrow \mathcal Z$, and prediction function $g_Y:\mathcal Z \rightarrow \mathcal Y$ on top of that representation such that,
\begin{align}
\label{nonlinear_model}
    Y = g_Y\circ\phi_Y(\mathbf X) + E = g_Y(\mathbf Z) + E,
\end{align}
where $g_Y := g(\cdot;\beta_0)$ is parameterized by the causal parameter $\beta_0$ and can be considered a linear map without loss of generality. $E$ stands for potential sources of mispecification and unexplained sources of variability, such as due to unobserved confounding and in general $\mathbf Z \notindep E$. For a given sample of data $(\mathbf z,y)$, optimal parameters $\hat\beta$ are often taken to minimize squared residuals, with $\hat\beta$ the solution to the normal equations: $\nabla_{\beta} g(\mathbf z;\hat\beta)y = \nabla_{\beta} g(\mathbf z;\hat\beta)g(z;\hat\beta)$, where $\nabla_{\beta} g(\mathbf z;\hat\beta)$ denotes the column vector of gradients of $g$ with respect to parameters $\beta$ evaluated at $\hat\beta$. Consider the Taylor expansion of $g(\mathbf z;\beta_0)$ around an estimate $\hat\beta$ sufficiently close to $\beta_0$, $g(\mathbf z;\beta_0) \approx g(\mathbf z;\hat\beta) + \nabla_{\beta} g(\mathbf z;\hat\beta)^T (\beta_0 - \hat\beta)$. Using the definition of $Y$ and replacing this approximation in our normal equations we find,
\begin{align}
\label{least_squares_consistency}
    \nabla_{\beta} g(\mathbf z;\hat\beta)\nabla_{\beta} g(\mathbf z;\hat\beta)^T(\hat\beta - \beta_0) + v = \nabla_{\beta} g(\mathbf z;\hat\beta) e,
\end{align}
where $v$ is a scaled disturbance term that includes the rest of the Taylor expansion of $g$ and is $\mathcal O((\hat\beta - \beta_0)^2)$ as $(\hat\beta - \beta_0)\rightarrow 0$. $e:= y - g(\mathbf z;\hat\beta)$ is the residual. $\hat \beta$ is consistent for the true $\beta_0$ if and only if $\nabla_{\beta} g(\mathbf z;\hat\beta) e \rightarrow 0$ in probability which is unlikely to be true in the presence of unobserved confounding due to the dependence between $\mathbf Z$ and $E$. Convergence to the causal solution in standard regression tasks can be expected only if $E$ is independent of $\mathbf X$ which is not the case in the presence of unobserved confounding and is one reason for the poor performance of empirical risk minimization in changing environments.

%We have access to observations of variables $(X_1,X_2,Y)$ in two training datasets, each dataset sampled with differing variances ($\sigma=1$ and $\sigma = 2$) from the following structural model $\mathbb F$,
%\begin{align*}
%    X_2 := -H + E_{X_2}, \quad Y := X_2 + 3H + E_{Y}, \quad X_1 := Y + X_2 + E_{X_1}
%\end{align*}
%where $E_{X_1}, E_{X_2}\sim\mathcal N(0,\sigma)$, $E_Y\sim\mathcal N(0,1)$ are exogenous variables. In a first scenario (leftmost plot) we consider all data (training and testing) to be generated \textbf{without} unobserved confounders, $H:=0$; and, in a second scenario (remaining plots) all data \textbf{with} unobserved confounders, $H:=E_H\sim\mathcal N(0,1)$.

However, even if the inner product is \textit{non-zero} $\nabla_{\beta} g(\mathbf z;\hat\beta) e \neq 0$, it does converge to a \textit{fixed unknown value} that is equal across environments under certain assumptions. 

\textbf{Proposition 1} (Derivative invariance). \textit{For any two environment distributions $P_i$ and $P_j$ generated under variations of an additive, invertible SCM as defined in Definition 2, it holds that the causal parameter $\beta_0$ satisfies,
\begin{align}
    \underset{(\mathbf x,y)\sim P_i}{\mathbb E}\nabla_{\beta} g(\mathbf z;\beta_0)(y - g(\mathbf z;\beta_0)) - \underset{(\mathbf x,y)\sim P_j}{\mathbb E}\nabla_{\beta} g(\mathbf z;\beta_0)(y - g(\mathbf z;\beta_0)) = 0.
\end{align}
A parameter $\hat\beta$ solution to,
\begin{align}
\label{optimal_beta}
    \underset{(\mathbf x,y)\sim P_i}{\mathbb E}\nabla_{\beta} g(\mathbf z;\beta)(y - g(\mathbf z;\beta)) - \underset{(\mathbf x,y)\sim P_j}{\mathbb E}\nabla_{\beta} g(\mathbf z;\beta)(y - g(\mathbf z;\beta)) = 0.
\end{align}
is consistent for the causal parameter $\beta_0$ if unique even in the presence of unobserved confounders.}

A few remarks are necessary concerning this statement and its implications for performance generalization.

\subsection{Remarks}

\begin{itemize}[leftmargin=*]
    \item The first remark is based on the observation that, up to a constant, each inner product in (\ref{optimal_beta}) is the gradient of the squared error with respect to $\beta$. This reveals that the optimal predictor, in the presence of unobserved confounding, is not one that produces minimum loss but one that produces a \textit{non-zero} loss gradient \textit{equal} across environments. Therefore, seeking minimum error solutions, even in the population case, produces estimators with \textit{necessarily} unstable correlations because the variability due to unobserved confounders is not explainable from observed data.
    
    Forcing gradients to be zero then \textit{forces} the learned model to utilize artifacts of the specific data collection process that are not related to the causal input-output relationship; and, for this reason, will not in general perform outside training data. 
    
    \item From eq. (\ref{optimal_beta}) we may pose a sequence of moment conditions for each pair of available environments. We may then seek solutions $\beta$ that make all of them small simultaneously. Solutions are unique if the set of moments is sufficient to identify $\beta_0$ exactly. We revisit our introductory example to illustrate that indeed this is the case. Table \ref{bias_main} shows the estimated coefficients of a linear model trained for minimum average error (ERM) and trained using the moment condition in eq. (\ref{optimal_beta}). As a contrast with related work, IRM \cite{arjovsky2019invariant} and REx \cite{krueger2020out} define two different invariances for causal solutions that are not valid in the presence of unobserved confounders. We discuss this in more detail in Appendix x.
    
    \begin{table}[H]
    \fontsize{8.5}{9.5}\selectfont
    \centering
    \begin{tabular}{ |p{3.7cm}|C{1.5cm}|C{1.5cm}|C{1.5cm}|C{1.5cm}|C{1.5cm}|  }
     \cline{2-6}
      \multicolumn{1}{c|}{} & \textbf{Truth} & \textbf{ERM} & \textbf{IRM} & \textbf{REx} & \textbf{Eq. (\ref{optimal_beta})}  \\
     \hline
     Est. coefficients for $(X_1,X_2)$ & (0, 1) & (0.91, -1.02) & (0.75, -0.76) & (0.14, 0.62) & (0.01, 0.95) \\
     \hline
    \end{tabular}
    \caption{Estimated parameters on data sampled with unobserved confounders ($\eta = 1$) from the motivating example.}
    \label{bias_main}
    \end{table}

   \item In practice, only a potentially infinite set of solutions may be identified with the moment conditions given in Proposition 1 with no performance guarantees for any individual solution, and no guarantees if assumptions on the data generating process fail to hold. And, even if accessible, we have seen that causal solutions may not always be desirable under moderate shifts in the distribution of covariates or under shifts in unobserved variables for example. It will be interesting to study a relaxation of strict causal invariances with the objective of learning solutions with a more desirable performance profile in different environments.
\end{itemize}

\section{A Robust Optimization Perspective}
\label{sec_gen}

In this section we motivate a relaxation of the ideas presented above using the language of robust optimization. For this purpose we consider a different kind of uncertainty set $\mathcal P$ in problems of the form of (\ref{robust_pop}) that specify an \textit{affine} combination of training losses. Let $\Delta_{\eta}:=\{\{\alpha_e:e\in\mathcal E\}: \alpha_e \geq -\eta, \sum_{e\in\mathcal E} \alpha_e = 1\}$ be a collection of scalars and consider the set of distributions defined by $\mathcal P := \{\sum_{e\in\mathcal E} \alpha_e P_e : \{\alpha_e:e\in\mathcal E\} \in\Delta_{\eta}\}$. $\eta \in \mathbb R$ defines the strength of the extrapolation, $\eta = 0$ corresponds to a convex hull of distributions\footnote{\cite{krueger2020out} have noted that in high-dimensional systems perturbed data is likely to occur at a new vertex not represented as a linear combination of training environments.} and for $\eta > 0$ the space of distributions defined by $\mathcal P$ is richer, going beyond what has been observed, amplifying the "strength" of manipulations that generated the observed training environments and explored in more detail in section \ref{sec_rob_inter}. 

Let $\mathcal L\left(f(\mathbf x),y \right) = \left(y - f(\mathbf x)\right)^2$ be the squared error for an input output pair $(\mathbf x,y)$ and model $f=g \circ \phi$, parameterized by $\beta$ and $\phi$. The following theorem presents a connection between invariances in loss derivatives and robustness to affine combinations of data distributions.

\textbf{Theorem 1} \textit{Let $\{P_e: e \in \mathcal E\}$ be a set of available environments indexed by $\mathcal E$. Further, assume the $\beta$ parameter space to be bounded $\|\beta\|_{\infty} \leq W$. Then, the following inequality holds,}
\begin{align*}
    \underset{\{\alpha_e\} \in \Delta_{\eta}}{\sup}\hspace{0.1cm} \sum_{e\in\mathcal E} \alpha_e \underset{(\mathbf x,y)\sim P_e}{\mathbb E} &\mathcal L\left(f(\mathbf x),y \right) \leq \underset{(\mathbf x,y)\sim P_e, e\sim \mathcal E}{\mathbb E}  \mathcal L\left(f(\mathbf x),y \right) \\
    &+ (1 + n\eta) \cdot 2W \cdot 
    \Big| \hspace{0.1cm} \underset{e\in \mathcal E, \|\beta\|_{\infty}<W}{\sup}\hspace{0.1cm}\underset{(\mathbf x,y)\sim P_e}{\mathbb E}  \nabla_{\beta}\mathcal L\left(f(\mathbf x),y \right) - \underset{(\mathbf x,y)\sim P_e, e\sim \mathcal E}{\mathbb E} \nabla_{\beta}\mathcal L\left(f(\mathbf x),y \right)\hspace{0.1cm} \Big |\\
     &+ (1 + n\eta) \cdot \Big|\hspace{.1cm}\underset{e\in \mathcal E}{\sup}\hspace{.1cm}\underset{y\sim P_e}{\mathbb E}  y^2 -\underset{y\sim P_e, e\sim \mathcal E}{\mathbb E} y^2 \Big|,
\end{align*}
\textit{where $n:= |\mathcal E|$ is the number of available environments and $e\sim\mathcal E$ loosely denotes sampling indices with equal probability from $\mathcal E$.}

This bound suggests a trade-off between prediction in-sample (first term) and invariances in loss derivatives (second term). The third term is the largest absolute deviation in the irreducible variance of the outcome with respect to its mean value across environments. A combination of them upper-bounds a robust optimization problem over affine combinations of training environments, and depending how much we choose to weight each objective (prediction versus invariance) we can expect solutions minimizing the right hand side of Theorem 1 to be more or less robust. Specifically, for $\eta = -1/n$ the $\{\alpha_e\}$ are fixed to $1/n$ and the objective reduces to ERM; for $\eta = 0$ this objective reduces to optimizing for the worst loss in a convex combination environments; $\eta \rightarrow \infty$ the objective forces solutions to have invariant derivatives across environments, it is robust at least to \textit{any} affine combination of environments and under certain conditions on the differences between environments converges to the causal solution.

\textbf{Remark on assumptions.} As long as new data distributions can be represented as affine combinations of training distributions, we can expect performance to be as least as good as that observed for the robust problem in Theorem 1, without necessarily requiring environments and underlying SCMs to be as strictly defined as in Definitions 1 and 2 (even though assuming the existence of a common SCM will allow us to understand the semantics of the expected robustness in terms of interventions in section \ref{sec_rob_inter}).

\subsection{Proposed objective} 
Our proposed learning objective is to guide the optimization of $\phi$ and $\beta$ towards solutions that approximately minimize the upperbound in Theorem 1,
\begin{align}
\label{robust_objective}
    \underset{\beta,\phi}{\text{minimize }}\underset{(\mathbf x,y)\sim P_e, e\sim \mathcal E}{\mathbb E} \mathcal L\left(f(\mathbf x),y \right) + \lambda\cdot \underset{e\sim \mathcal E}{\text{Var}}\left(\underset{(\mathbf x,y)\sim P_e}{\mathbb E}\nabla_{\beta}\mathcal L\left(f(\mathbf x),y \right)\right),
\end{align}
%Theorem 1 makes it clear that extrapolation guarantees are entirely defined in terms of the available environments. Their properties dictate the distributions we can expect to be robust to, which is a limitation in practice if these are poorly understood. 

where $\lambda \geq 0$. The variance is used instead of the absolute deviation for efficiency of optimization following the arguments of \cite{krueger2020out} and instead of fixing $\beta$ to the largest derivatives $\nabla_{\beta}\mathcal L\left(f(\mathbf x),y \right)$ we optimize it using both terms in the objective. We call this problem Derivative Invariant Risk Minimization (DIRM). 

\textbf{Remark on related work.} This objective bears similarities with REx \cite{krueger2020out}, a recent proposal for domain generalization that encourages equality between environment losses, loosely written minimizing: $\mathbb E \mathcal L\left(f(\mathbf x),y \right) + \lambda\cdot \text{Var}\left(\mathbb E\mathcal L\left(f(\mathbf x),y \right)\right)$, instead of equality between derivatives of environment losses. The interpretation of causality of certain solutions is lost in the presence of unobserved confounders as we do not expect losses to be equal across environments at optimum due to potential changes in the irreducible error across environments or due to interventions on target variables themselves.

%We would expect data in these environments to satisfy some regularities, for instance the causal coefficients in (\ref{full_model}) should remain the same and in particular the strength of confounding, that is, the correlation between unobserved confounders $Z$ and observed covariates $\mathbf X$ and thus $\nabla_{\beta} f(z;\hat \beta)^T E$ should remain the same. In this case, with data from two environments $\mathbf X_1, \mathbf X_2$,
%\begin{align}
%    (\mathbf X_1^T\mathbf X_1 - \mathbf X_2^T\mathbf X_2) (\beta - \hat\beta) = \mathbf X_1^T E_1 - \mathbf X_2^T E_2 \rightarrow 0
%\end{align}
%in probability, where $\hat\beta$ is such that $\mathbf (\mathbf X_1^T\mathbf X_1 - \mathbf X_2^T\mathbf X_2) \hat\beta = \mathbf X_1^T \mathbf y_1 - \mathbf X_2^T \mathbf y_2$ and is consistent for the causal effect $\beta$. In model (\ref{full_model}), constant confounding means $\alpha M \text{Cov}(x)$ is constant across environments, but data generating mechanisms may differ otherwise and must be sufficiently different to uniquely identify $\beta$. In fact, $\hat\beta$ is unique in this model if and only if $(\mathbf X_1^T\mathbf X_1 - \mathbf X_2^T\mathbf X_2)$ is full rank in expectation which holds in the population case if and only if environment-specific interventions on all dimensions of $E_x$ led to the different observed environments (see Theorem 2 in \cite{rothenhausler2019causal}).

\subsection{Robustness in terms of interventions}
\label{sec_rob_inter}
In this section we give a causal perspective on the robustness achieved by our objective in (\ref{robust_objective}). As is apparent in Theorem 1, performance guarantees on data from a new environment depend on the relationship of new distributions with those observed during training.

Let $f_{\lambda \rightarrow \infty}$ minimize $\mathcal L$ among all functions that satisfy all pairs of moment conditions defined in (\ref{optimal_beta}); that is, a solution to our proposed objective in (\ref{robust_objective}) with $\lambda\rightarrow\infty$. At optimality, it holds that gradients evaluated at this solution are equal across environments. As a consequence of Theorem 1, the loss evaluated at this solution with respect to \textit{any} affine combination of environments is bounded by the average loss computed in-sample, denoted $L$,
\begin{align}
    \sum_{e\in\mathcal E} \alpha_e \underset{(\mathbf x,y)\sim P_e}{\mathbb E} \mathcal L\left(f(\mathbf x),y \right) \leq L, \qquad\text{for any set } \{\alpha_e:e\in\mathcal E\} \in \Delta_{\eta}.
\end{align}
Solutions $f_{\lambda \rightarrow \infty}$ can be interpreted as enjoying a form of data-driven predictive stability: having stable performance across the range of distributions generated from an underlying SCM intervened on in the same "direction" as those observed during training. 

Consider an example to describe what we mean by the direction of perturbations. Consider distributions $P$ of a univariate random variable $X$ given by affine combinations of training distributions $P_0$ with mean $0$ and $P_1$ which is perturbed to have mean $1$ so that, using our notation, $\mathbb E_P X = \alpha_0\mathbb E_{P_0} X + \alpha_1\mathbb E_{P_1}X$, $\alpha_0=1-\alpha_1\geq -\eta$. $\mathbb E_P X\in[-\eta,\eta]$ and thus we may expect DIRM to be robust to distributions subject to interventions of magnitude $\pm\eta$ on $ X$ and any magnitude in the limit $\eta\rightarrow\infty$ (or equivalently $\lambda\rightarrow\infty$). 

Note that the "diversity" of training environments has a large influence on whether we can interpret solutions to be causal or robust; for instance, with equal means in $P_0$ and $P_1$ affine combinations would not extrapolate to interventions in the mean of $X$. The "type" of shift to be observed in new data has to be present among training environment:, shifts must occur in the same direction. However, contrary to causal solutions defined from a robust optimization perspective to have stable performance on distribution shifts in observed variables, see e.g. \cite{meinshausen2018causality}, $f_{\lambda \rightarrow \infty}$ can be expected to have stable performance on observed, unobserved or even target variables as long as that shift is present in training environments.

\begin{figure}[H]
\captionsetup{skip=1pt}
\centering
\includegraphics[width=0.8\textwidth]{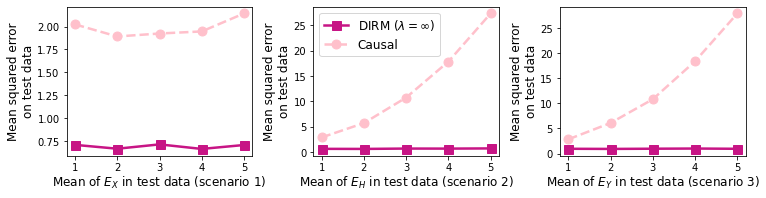}
\caption{Performance stability to general shifts when present in training environments.}
\label{stability}
\end{figure}

We illustrate this using our motivating example by specifying three scenarios: \textbf{1)} training environments sampled with $\mathbb E_{P_0} E_{X_1} = 0$ and $E_{P_1} E_{X_1} = 1$, \textbf{2)} training environments sampled with $\mathbb E_{P_0}H = 0$ and $E_{P_1}H = 1$, and \textbf{3)} training environments sampled with $\mathbb E_{P_0}E_Y = 0$ and $E_{P_1}E_Y = 1$, everything else being equal i.e. ($\sigma^2 := 1, \eta=1$). In Figure \ref{stability} we plot mean squared errors on data sampled by increasing the shift present in each one of the three scenarios, e.g. 1) $\mathbb E_{P_{new}}E_{X_1} \in [0,5]$, 2) $\mathbb E_{P_{new}} E_H \in [0,5]$, and so on. In all cases, we see that the performance of $f_{\lambda \rightarrow \infty}$ is stable to increasing perturbations in the system while causal solutions are only stable to perturbations in $\mathbf E_X$. 

\subsection{Stability of certain optimal solutions} 
\label{stability_section}
A special case may also be considered when the underlying system of variables and the available environments allow for optimal solutions $f_{\lambda \rightarrow \infty}$ and $f_{\lambda = 0}$ to coincide. In this case, the learned representation $\phi(\mathbf x)$ results in a predictor $g$ optimal on average \textit{and} simultaneously with equal gradient in each environment, thus,
\begin{align*}
    |\underset{(\mathbf x,y)\sim P_e}{\mathbb E}\nabla_{\beta}\mathcal L\left(f(\mathbf x),y \right)| = 0, \qquad \text{for all } e\in\mathcal E.
\end{align*}
For this representation $\phi$, it follows that optimal solutions $f$ learned on any new dataset sampled from an affine combination of training distributions coincides with this special solution. This gives us a sense of \textit{reproducibility of learning} which we will explore in the experimental section \ref{sec_reproducibility}: if a specific feature is significant for predictions on the whole range of $\lambda$ with the available data then it will likely be significant on new (related) data. 

\textbf{Remark on related work.} The special case where all solutions in the range of hyperparameter $\lambda$ agree has important parallels with Invariant Risk Minimization (IRM) \cite{arjovsky2019invariant}. The authors proposed a learning objective enforcing representations of data with minimum error on average and across environments, such that at optimum $\mathbb E_{P_i} Y|\hat\phi(\mathbf X) = \mathbb E_{P_j} Y|\hat\phi(\mathbf X)$ for any pair $(i,j)\in\mathcal E$. Without unobserved confounding, our proposal and IRM agree as an optimal causal predictor (with zero gradient) across all environments can be achieved. But, with unobserved confounding, minimum error causal solutions, with zero gradients in all environments, are not possible by design and IRM converges on solutions that necessarily exploit confounded dependencies that are not guaranteed to generalize from a causal perspective. For example, under an SCM defined by additive unobserved confounding $H$, e.g. $Y:= g\circ\phi(\mathbf X) + H + E_Y$, for the causal solution $g$ and $\phi$, $\mathbb E_{P_i} Y|\phi(\mathbf X) = g\circ\phi(\mathbf X) + \mathbb E_{P_i} H|\phi(\mathbf X) \neq  g\circ\phi(\mathbf X) + \mathbb E_{P_j} H|\phi(\mathbf X) = \mathbb E_{P_j} Y|\phi(\mathbf X)$ if the means of $H|\phi(\mathbf X)$ in $P_i$ and $P_j$ differ. The sought invariance then does not hold for the causal solution in general.

\section{Related work}

% https://arxiv.org/ftp/arxiv/papers/1205/1205.2640.pdf for related work on unobserved confounding
%Distributional shifts arise in many guises across topics in causal inference, machine learning and optimization. In this section we explore some of the emerging connections between these fields. 

There has been a growing interest in interpreting shifts in distribution to fundamentally arise from interventions in the causal mechanisms of data. Peters et al. \cite{peters2016causal} were among the first to exploit this link for causal inference by looking for invariant dependencies across environments. Invariant solutions, as a result of this connection, may be interpreted also as robust to certain interventions, and a large amount of recent work has explored learning invariances in various problem settings from a causal and distributionally robust optimization perspective, see e.g. \cite{arjovsky2019invariant,rothenhausler2019causal,krueger2020out,gimenez2020identifying, meinshausen2018causality, jin2020enforcing, ahuja2020invariant, koyama2020out, shi2021gradient, rothenhausler2019causal, rothenhausler2018anchor,meinshausen2015maximin}. 

From a conceptual perspective, DIRM is closest to Anchor Regression \cite{rothenhausler2018anchor} which is also applicable to unobserved confounding and where the authors make a similar argument combining standard losses with a causal objective related to instrumental variable regression. In this work, we extend this analysis to non-linear structural causal models, derive a different causal objective that involves loss derivatives instead of instrumental variables and therefore also propose a different study of distributionally robust performance guarantees. In particular, DIRM does not reduce to Anchor Regression in linear SCMs.

From an algorithmic perspective, DIRM is close to equivalent to the objectives proposed by \cite{koyama2020out} and \cite{shi2021gradient}. Both explicitly use loss derivatives with respect to model parameters to regularize ERM solutions. \cite{koyama2020out} re-formulate the out-of-sample generalization problem as finding a predictor that minimizes the worst-case mutual information between representations $\phi(\mathbf X)$ and outcome $Y$ across a set of environments subject to the requirement that $Y\indep\epsilon|\phi(\mathbf X)$, where $\epsilon$ is an indicator for the environment, without however defining precisely what the differences between environments are or how this objective relates to robustness in practice. \cite{shi2021gradient} instead take an optimization approach and propose to optimize for minimum average loss while maximizing the inner product between gradients from different environments such that each parameter update is encouraged to simultaneously improve predictions in all environments. In contrast our contribution is on the derivation of such an objective from a structural model-based perspective and studying the implications for generalization from a causal and distributionally robust perspective.

%It includes as special cases, for instance, problems in domain adaptation, see e.g. \cite{ben2009robust,kuhn2019wasserstein,duchi2019distributionally,sinha2017certifying,wozabal2012framework,abadeh2015distributionally,duchi2018learning} and covariate shift, see e.g. \cite{bickel2009discriminative,duchi2019distributionally}.

% https://arxiv.org/pdf/1808.05541.pdf look at example 1 for a simple proof that ICP does not hold with unobserved variables.

% https://arxiv.org/pdf/1803.00810.pdf Very good paper on unobserved confounding, use for introduction and examples. 

% file:///C:/Users/abellot/Downloads/make-01-00019-v2.pdf Good related work

% https://arxiv.org/pdf/2001.06208.pdf Causal models for dynamical systems

% https://simons.berkeley.edu/sites/default/files/docs/10799/buhlmann-simons.pdf can still achieve “shift invariance” of residuals: a non-trivial fact is: (Y − Xb) is “shift-invariant” iff A uncorrelated with (Y − Xb) thus, we want to encourage orthogonality of A with the residuals. It is quite different from classical statistical robustness: robust stats: - downweight outliers to “approach” the reference distr. - aims (primarily) for training sample robustness I anchor/invariance: make use of/exploit the perturbations to inspect stability and hence “robustify” against against adversarial future scenarios/test samples

\section{Experiments}

In this section, we conduct an analysis of generalization performance on shifted image, speech and tabular data from the medical domain. We make comparisons on domain generalization benchmarks including VLCS \cite{fang2013unbiased}, PACS \cite{li2017deeper} and Office-Home \cite{venkateswara2017deep} using the DomainBed platform \cite{gulrajani2020search} in the Appendix. All experimental details are standardized across experiments and algorithms, i.e. equal network architectures and hyperparameter optimization techniques. All specifications can be found in the Appendix. 

Data linkages, electronic health records, and bio-repositories, are increasingly being collected to inform medical practice. As a result, also prediction models derived from healthcare data are being put forward as potentially revolutionizing decision-making in hospitals. Recent studies \cite{cabitza2017unintended,venugopalan2019s},
however, suggest that their performance may reflect not only their ability to identify disease-specific
features, but also their ability to exploit spurious correlations due to unobserved confounding (such as
varying data collection practices): a major challenge for the reliability of decision support systems.

In our comparisons we consider the following baseline algorithms:
\begin{itemize}[leftmargin=*, itemsep=0pt]
    \item Empirical Risk Minimization (\textbf{ERM}) that optimizes for minimum loss agnostic of data source.
    \item Group Distributionally Robust Optimization (\textbf{DRO}) \cite{sagawa2019distributionally} that optimizes for minimum loss across the worst convex mixture of training environments.
    \item Domain Adversarial Neural Networks (\textbf{DANN}) \cite{ganin2016domain} that use domain adversarial training to facilitate transfer by augmenting the neural network architecture with an additional domain classifier to enforce the distribution of $\phi(X)$ to be the same across training environments.
    \item Invariant Risk Minimization (\textbf{IRM})  \cite{arjovsky2019invariant} that regularizes ERM ensuring representations $\phi(X)$ be optimal in every observed environment.
    \item Risk Extrapolation (\textbf{REx}) \cite{krueger2020out} that regularizes for equality in environment losses instead of considering their derivatives.
\end{itemize}
%Performance results are given in Table \ref{perf}. 

\begin{table*}[t]
\fontsize{9.5}{9.5}\selectfont
\centering
\begin{tabular}{ |p{1.2cm}|C{1.6cm}|C{1.6cm}||C{1.6cm}|C{1.6cm}||C{1.6cm}|C{1.6cm}|  }

 \cline{2-7}
   \multicolumn{1}{c|}{} & \multicolumn{2}{c||}{\textbf{Pneumonia Prediction}} &   \multicolumn{2}{c||}{\textbf{Parkinson Prediction}}&\multicolumn{2}{c|}{\textbf{Survival Prediction}}\\
 \cline{2-7}
   \multicolumn{1}{c|}{} & \textbf{Training} & \textbf{Testing} & \textbf{Training} & \textbf{Testing}&\textbf{Training} & \textbf{Testing}\\
 \hline
 ERM &  91.6 ($\pm$ .7)  & 52.7 ($\pm$ 1)  &  95.5 ($\pm$ .5)  & 62.8 ($\pm$ 1) & 93.2 ($\pm$ .4) & 75.4  ($\pm$ .9)\\
 \hline
 DRO  &  91.2 ($\pm$ .5)  & 53.0 ($\pm$ .6)  &  94.0 ($\pm$ .3) & 69.9 ($\pm$ 2)& 90.4 ($\pm$ .4) & 75.2  ($\pm$ .8) \\
 \hline
 DANN  &  91.3 ($\pm$ 1)  & 57.7 ($\pm$ 2)  &  91.6 ($\pm$ 2) & 51.4 ($\pm$ 5) & 89.0 ($\pm$ .8) & 73.8  ($\pm$ .9)\\
 \hline
 IRM &  89.3  ($\pm$ 1)  & 58.6 ($\pm$ 2)  &  93.7 ($\pm$ 1) & 71.4 ($\pm$ 2)& 91.7 ($\pm$ .6) & 75.6  ($\pm$ .8)\\
 \hline 
 REx &  87.6  ($\pm$ 1)  & 57.7 ($\pm$ 2)  &  92.1 ($\pm$ 1) & 72.5 ($\pm$ 2)& 91.1 ($\pm$ .5) & 75.1  ($\pm$ .9)\\
 \hline
 \textbf{DIRM} &  84.4 ($\pm$ 1)  & 63.1 ($\pm$ 3)  &  93.0  ($\pm$ 2)& 72.4 ($\pm$ 2) & 91.2 ($\pm$ .6) & 77.6  ($\pm$ 1) \\
 \hline
\end{tabular}
\caption{Accuracy of predictions in percentages ($\%$). Uncertainty intervals are standard deviations. All datasets are approximately balanced, $50\%$ performance is as good as random guessing.}
\label{perf}
\end{table*}

\subsection{Diagnosis of Pneumonia with Chest X-ray Data}
In this section, we attempt to replicate the study in \cite{zech2018confounding}. The authors observed a tendency of image models towards exploiting spurious correlations for the diagnosis on pneumonia from patient Chest X-rays that do not reproduce outside of training data. We use publicly available data from the National Institutes of Health (NIH) \cite{wang2017chestx} and the Guangzhou Women and Children’s Medical Center (GMC) \cite{kermany2018identifying}. Differences in distribution are manifest, and can be seen for example in the top edge of mean pneumonia-diagnosed X-rays shown in Figure \ref{x_ray}. In this experiment, we exploit this (spurious) pathology correlation to demonstrate the need for solutions robust to changes in site-specific features.

\begin{minipage}{.6\textwidth}
\textbf{Experiment design.}  We construct two training sets that will serve as training environments. In each environment, $90\%$ and $80\%$ of pneumonia-diagnosed patients were drawn from the NIH dataset and the remaining $10\%$ and $20\%$ of the pneumonia-diagnosed patients were drawn from the GMC dataset.  The reverse logic ($10\%$ NIH / $90\%$ GMC split) was followed for the test set. This encourages algorithms to use NIH-specific correlations for prediction during training which are no expected to extrapolate during testing.
\end{minipage}
\hfill
\begin{minipage}{.32\textwidth}
\begin{figure}[H]
\vspace{-0.4cm}
\captionsetup{skip=5pt}
\centering
\includegraphics[width=1\textwidth]{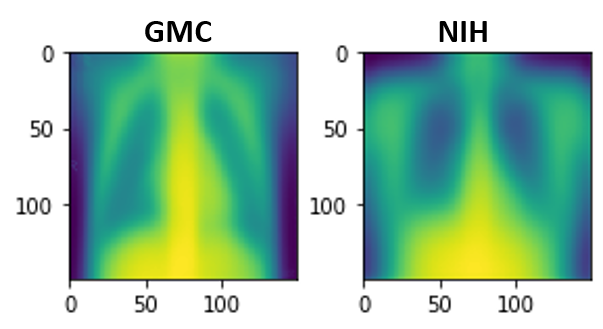}
\caption{Mean pneumonia X-ray.}
\label{x_ray}
\end{figure}
\end{minipage}

Our results (Table \ref{perf}) show that DIRM significantly outperforms, suggesting that the proposed invariance guides the algorithm towards better solutions in the case of changes due to unobserved factors.

%In all splits the classes and site distributions were always balanced, making it tempting for the classifier to use a site-specific feature when predicting the class in the presence of the spurious site-pathology correlation. 1006 samples are available in each training environment and 574 samples in the testing set. 

%Our results can be found in Table \ref{xray_perf}. The performance dynamics are consistent with our discussion so far and the trade-off between performance in-sample and out-of-sample is manifest. We hypothesize that IRM, while improving upon ERM, lags behind our approach in terms of out-of-sample performance because of potential unobserved confounders. We explore this point further in our second experiment next.

\subsection{Diagnosis of Parkinson's Disease with Speech}
Parkinson's disease is a progressive nervous system disorder that affects movement. Symptoms start gradually, sometimes starting with a barely noticeable tremor in a patient's voice. This section investigates the performance of predictive models for the detection of Parskinson's disease, trained on voice recordings of vowels, numbers and individual words and tested on vowel recordings of unseen patients. 

\textbf{Experiment design.} We used the UCI Parkinson Speech Dataset with given training and testing splits \cite{sakar2013collection}. Even though the distributions of features will differ in different types of recordings and patients, we would expect the underlying patterns in speech to reproduce across different samples. However, this is not the case for correlations learned with baseline training paradigms (Table \ref{perf}). This suggests that spurious correlations due to the specific type of recording (e.g. different vowels or numbers), or even chance associations emphasized due to low sample sizes (120 examples), may be responsible for poor generalization performance. Our results show that correcting for spurious differences between recording types (DIRM, IRM, REx) can improve performance substantially over ERM although the gain of DIRM over competing methods is less pronounced.

\subsection{Survival Prediction with Health Records}
This section investigates whether predictive models transfer across data from different medical studies \cite{meta2012survival,bellot2019boosting,bellot2018boosted}, all containing patients that experienced heart failure. The problem is to predict survival within 3 years of experiencing heart failure from a total of 33 demographic variables. We introduce a twist however, explicitly introducing unobserved confounding by omitting certain predictive variables. The objective is to test performance on new studies with \textit{shifted} distributions, while knowing that these occur predominantly due to variability in unobserved variables. 

\textbf{Experiment design.} Confounded data is constructed by omitting a patient's age from the data, found in a preliminary correlation analysis to be associated with the outcome as well as other significant predictors such as blood pressure and body mass index (that is, it confounds the association between blood pressure, body mass index, and survival). This example explicitly introduces unobserved confounding, but this scenario is expected in many other scenarios and across application domains. For instance, such a shift might occur if a prediction model is taken to patients in a different hospital or country than it was trained on. Often distribution of very relevant variables (e.g. socio-economic status, ethnicity, diet, etc.) will differ even though this information is rarely recorded in the data. We consider the 5 studies in MAGGIC of over 500 patients with balanced death rates. Performance results are averages over 5 experiments, in each case, one study is used for testing and the remaining four are used for training. DIRM's performance in this case is competitive with methods which serves to confirm the desirable performance profile of DIRM.

\subsubsection{Reproducibility of variable selection} 
\label{sec_reproducibility}

Prediction algorithms are often use to infer influential features in outcome prediction. It is important that this inference be consistent across environments even if perturbed or shifted in some variables. Healthcare is challenging in this respect because patient heterogeneity is high. We showed in section \ref{stability_section} that in the event that the optimal predictor is invariant as a function of $\lambda\in[0,\infty)$, optimal predictors estimated in \textit{every} new dataset in the span of observed distributions, should be \textit{stable}. We test this aspect in this section, considering a form of diluted stability for feature selection ($\lambda\in[0,1]$ instead of $\lambda\in[0,\infty)$).

\begin{minipage}{.6\textwidth}
\textbf{Experiment design.}  For a single layer network, we consider significant those covariates with estimated parameters bounded away from zero in all solutions in the range $\lambda\in[0,1]$. Comparisons are made with ERM (conventional logistic regression) and both methods are trained separately on 100 different random pairs of the 33 MAGGIC studies, that is 100 different environments on which algorithms may give different relevant features. Figure \ref{maggic} shows how many features (among the top 10 discovered features) in each of the 100 experiments intersect. For instance, we have that $6$ features intersecting across $80/100$ runs for DIRM while only $4$ for ERM (approximately). DIRM thus recovers influential features more consistently than ERM.
\end{minipage}
\hfill
\begin{minipage}{.37\textwidth}
\begin{figure}[H]
\vspace{-1.5em}
\captionsetup{font=small,skip=0pt}
\centering
\includegraphics[width=0.6\textwidth]{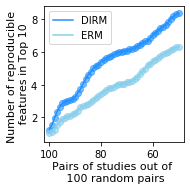}
\caption{Reproducibility of variable selection.}
\label{maggic}
\end{figure}
\end{minipage}

\section{Conclusions}
We have studied the problem of out-of-sample generalization from a new perspective, grounded in the underlying causal mechanism generating new data that may arise from shifts in observed, unobserved or target variables. Our proposal is a new objective, DIRM, that is provably robust to certain shifts in distribution, and is informed by new statistical invariances in the presence of unobserved confounders. Our experiments show that we may expect better generalization performance and also better reproducibility of influential features in problems of variable selection. A limitation of DIRM is that robustness guarantees crucially depend on the (unobserved) properties of available data: DIRM generally does not guarantee protection against unsuspected events. For example, in Theorem 1, the supremum contains distributions that lie in the affine combination of training environments, as opposed to arbitrary distributions.

\section*{Acknowledgements}
This work was supported by the Alan Turing Institute under the EPSRC grant EP/N510129/1, the ONR and the NSF grants number 1462245 and number 1533983.

\bibliography{bibliography}
\bibliographystyle{plain}

\newpage
\appendix
{\Large \textbf{Appendix}}
\\\\
This Appendix is outlined as follows.
\begin{itemize}[leftmargin=*]
    \item Section \ref{sec_related_work} discusses additional related work.
    \item Section \ref{bias_experiements} analyses the causal interpretation one may give to DIRM if conditions for causality are satisfied and makes additional comparisons with IRM and its invariance principle.
    \item Section \ref{proofs} provides proofs for the statements made in the main body of this paper.
    \begin{itemize}
        \item Section \ref{invariance_proof} gives further context to the assumptions and interventions needed for causality and proves Proposition 1.
        \item Section \ref{theorem_proof} proves Theorem 1.
    \end{itemize}
    \item Section \ref{sec_domainbed} gives performance comparisons on benchmarks datasets VLCS, PACS and Office-Home.
    \item Section \ref{experimental_details} gives details on experiments, implementation and data.
    
\end{itemize}

\newpage
\section{Additional related work}
\label{sec_related_work}

\textbf{Note on terminology.} In domain generalization, the training data is sampled from one or many source environments, while the test data is sampled from a new target domain. In contrast to domain adaptation, the learner does not have access to any data from the target domain (labeled or unlabeled) during training time. In this paper we are interested in the scenario where multiple source environments are available, and the domain where the data comes from is known but we have no information on the source of the potential shift in test domain nor any knowledge of the underlying causal graph or difference between training environments, even though as we have explained in the main body of this paper, knowledge of what variables have been intervened on does tell us the kind of generalization power that can be expected with DIRM. 

\textbf{Related work on the topics of invariance, causality and robustness. }There are a number of related work that include aspects of this formalism and ideas relating to invariance and causality for domain generalization. Besides the related work mentioned in the main body of this work, an important line of research starts from a known causal graph and known differences between environments to study what queries are identifiable in a given target domain; a field also known as transportability originating in \cite{pearl2011transportability,bareinboim2012transportability}. With this formalism, \cite{subbaswamy2019preventing,subbaswamy2018counterfactual} use this form of known heterogeneity between environments to remove unstable paths from the conditioning set of a prediction model, having explicit domain generalization guarantees. Similarly, \cite{magliacane2017domain} learn the causal graph with independence tests before learning features whose dependence with the target is invariant across environments (and did so in a domain adaptation problem formulation.

Notions of invariance have been found useful in the broader field of domain generalization without necessarily referring explicitly to an underlying causal model. For instance, recent work has included the use data augmentation \cite{volpi2018generalizing,shankar2018generalizing}, meta-learning to simulate domain shift \cite{li2018learning,zhang2020adaptive}, contrastive learning \cite{kim2021selfreg}, adversarial learning of representations invariant to the environment \cite{ganin2016domain, albuquerque2019adversarial}, and with applications in structured medical domains \cite{jin2020enforcing}. We note also recent papers that have generalized the principle of invariant risk minimization in various ways, some examples are \cite{lu2021nonlinear} and \cite{ahuja2020invariant}.

In a different line of research, instead of appealing explicitly to invariances between environments, many authors have proposed to solve directly a worst-case optimization problem (\ref{robust_pop}). One popular approach is to define $\mathcal P$ as a ball around the empirical distribution $\hat P$, for example using $f$-divergences or Wasserstein balls of a defined radius, see e.g. \cite{kuhn2019wasserstein,duchi2016statistics,duchi2019distributionally,sinha2017certifying,wozabal2012framework,abadeh2015distributionally,duchi2018learning}. These are general and multiple environments are not required, but this also means that sets are defined agnostic to the geometry of plausible shifted distributions, and may therefore lead to solutions, when tractable, that are overly conservative or do not satisfy generalization requirements \cite{duchi2019distributionally}.

\newpage
\section{Violation of invariances in the presence of unobserved confounders}
\label{bias_experiements}

So far, we have mainly considered predictive performance under different data distributions without explicitly considering the causal interpretation that can be given to DIRM. In this section we give more context to our remark in section \ref{sec_2} comparing parameter estimates of different invariant objectives (see Table \ref{bias_main}).

Recall the data generating mechanism of our motivating example. We assume access to observations of variables $(X_1,X_2,Y)$ in two training datasets, each dataset sampled with different variances ($\sigma^2=1$ and $\sigma^2 = 2$) from the following structural model,
\begin{align*}
    X_2 := -\eta H + (1-\eta)E_{X_2}, \quad Y := X_2 + \eta H + (1-\eta)E_{Y},\quad X_1 := Y + X_2 + E_{X_1}.
\end{align*}
$E_{X_1}, E_{X_2}\sim\mathcal N(0,\sigma^2)$, $E_Y\sim\mathcal N(0,1)$ are independent error terms and $H\sim\mathcal N(0,1)$ is an unobserved confounder whose influence is determined by $\eta\in\{0,1\}$.

\textbf{Results.} Given the above two training datasets, we inspect the weights learned in a simple one layer feed-forward neural network to determine exactly whether unobserved confounding induces a given learning paradigm to exploit spurious correlations and to what extent. 

In principle, causal solutions are recoverable with DIRM (with $\lambda\rightarrow\infty$) because we do observe environments with shifts in both $p(X_1)$ and $p(X_2)$ (irrespective of the presence or not of unobserved confounders). That is, conditions for Lemma 1 are satisfied and solutions are unique. We demonstrate this fact empirically in Table \ref{bias}. We see that this holds approximately for the proposed objective with estimated coefficients $(0.01, 0.95)$ for $(X_1,X_2)$ close to the true causal coefficients $(0,1)$. In contrast, ERM returns biased coefficients and so does IRM, which highlights the fact that enforcing minimum gradients on average (ERM) or simultaneously across environments (the regularization proposed by IRM) is not appropriate to recover causal coefficients in the presence of unobserved confounders. If, however, no unobserved confounders exist in the system being modelled ($H:=0$ in the data generating mechanism) our objective and IRM are equivalent in the limit, and estimated parameters coincide with the causal solution approximately. This experiment is given in Table \ref{nobias}.

\begin{table}[H]
\fontsize{8.5}{9.5}\selectfont
\centering
\begin{tabular}{ |p{3.7cm}|C{0.8cm}|C{1.5cm}|C{2cm}|C{2.2cm}|  }
 \cline{2-5}
  \multicolumn{1}{c|}{} & \textbf{Truth} & \textbf{ERM} & \textbf{IRM} $(\lambda\rightarrow\infty)$ & \textbf{DIRM} $(\lambda\rightarrow\infty)$  \\
 \hline
 Est. coefficients for $(X_1,X_2)$ & (0, 1) & (0.91, -1.02) & (0.75, -0.76) & (0.01, 0.95) \\
 \hline
\end{tabular}
\caption{Estimated parameters on data sampled \textbf{with} unobserved confounders. Both ERM and IRM give biased results due to the presence of unobserved confounding which highlights the fact that the invariance sought by IRM is not sufficient for consistency in this setting.}
\label{bias}
\end{table}

\begin{table}[H]
\fontsize{8.5}{9.5}\selectfont
\centering
\begin{tabular}{ |p{3.7cm}|C{0.8cm}|C{1.5cm}|C{2cm}|C{2.2cm}|  }
 \cline{2-5}
  \multicolumn{1}{c|}{} & \textbf{Truth} & \textbf{ERM} & \textbf{IRM} $(\lambda\rightarrow\infty)$ & \textbf{DIRM} $(\lambda\rightarrow\infty)$  \\
 \hline
 Est. coefficients for $(X_1,X_2)$ & (0, 1) & (0.5, -0.6) & (0.01, 0.98) & (0.02, 0.96) \\
 \hline
\end{tabular}
\caption{Estimated parameters on data sampled \textbf{without} unobserved confounders. Only ERM gives biased results since the conditions for consistency are satisfied for IRM and DIRM.}
\label{nobias}
\end{table}

\newpage

\section{Technical results}
\label{proofs}
This section provides a more complete discussion of the assumptions and justification statements relating to causality, Proposition 1, and the proof of Theorem 1.

\subsection{Proof Proposition 1}
\label{invariance_proof}

We restate the proposition for convenience.

\textbf{Proposition 1} (Derivative invariance). \textit{For any two environment distributions $P_i$ and $P_j$ generated under variations of an additive, invertible SCM as defined in Definition 2, it holds that the causal parameter $\beta_0$ satisfies,
\begin{align}
    \underset{(\mathbf x,y)\sim P_i}{\mathbb E}\nabla_{\beta} g(\mathbf z;\beta_0)(y - g(\mathbf z;\beta_0)) - \underset{(\mathbf x,y)\sim P_j}{\mathbb E}\nabla_{\beta} g(\mathbf z;\beta_0)(y - g(\mathbf z;\beta_0)) = 0.
\end{align}
A parameter $\hat\beta$ solution to,
\begin{align}
\label{optimal_beta}
    \underset{(\mathbf x,y)\sim P_i}{\mathbb E}\nabla_{\beta} g(\mathbf z;\beta)(y - g(\mathbf z;\beta)) - \underset{(\mathbf x,y)\sim P_j}{\mathbb E}\nabla_{\beta} g(\mathbf z;\beta)(y - g(\mathbf z;\beta)) = 0.
\end{align}
is consistent for the causal parameter $\beta_0$ if unique even in the presence of unobserved confounders.}

\textit{Proof.} An additive, invertible SCM $(\mathbf V, \mathbf U, \mathbf F, P)$ for $\mathbf V=(\mathbf X, \mathbf H, Y)$ and $\mathbf U=(\mathbf U_{\mathbf X}, \mathbf U_{\mathbf H}, U_{Y})$ may be written as follows,
\begin{align}
\label{additive_model}
    Y := f_1(\mathbf X) + f_2(\mathbf H) + U_{Y}, \quad \mathbf X := f_3(\mathbf X) + f_4(\mathbf H) + \mathbf U_{\mathbf X}, \quad \mathbf H := \mathbf U_{\mathbf H},
\end{align}
where $\mathbf H$ are assumed to stand for unobserved confounders. Under our definition of environments only $P(\mathbf U_{\mathbf X})$ varies between different datasets. We can rearrange the above to get,
\begin{align*}
    \mathbf X = (I - f_3)^{-1} f_4(\mathbf H) + (I - f_3)^{-1} \mathbf U_{\mathbf X} = (I - f_3)^{-1} f_4(\mathbf U_{\mathbf H}) + (I - f_3)^{-1} \mathbf U_{\mathbf X},
\end{align*} 
and therefore, with the parameterization assumed in the main body of this paper: $f_1(\mathbf X) := g \circ \phi (\mathbf X) = g(\mathbf Z, \beta_0)$ we have,
\begin{align*}
    \nabla_{\beta} g(\mathbf Z;\beta_0)(y - g(\mathbf Z;\beta_0)) = \nabla_{\beta}f_1(\mathbf X)(Y-f_1(\mathbf X)) = \big(\nabla_{\beta}f_1 (I - f_3)^{-1} f_4(\mathbf U_{\mathbf H}) + \nabla_{\beta}f_1 (I - f_3)^{-1} \mathbf U_{\mathbf X}\big) \cdot (f_2(\mathbf U_{\mathbf H}) + U_Y),
\end{align*}
which is a product of functions involving $\mathbf U_{\mathbf H}$ in one term, $\mathbf U_{\mathbf H}$ and $U_Y$ in another term, $\mathbf U_{\mathbf X}$ and $\mathbf U_{\mathbf H}$ in another term, and $\mathbf U_{\mathbf X}$ and $U_Y$ in the last term. Since $(\mathbf U_{\mathbf X}, \mathbf U_{\mathbf H}, U_{Y})$ are mutually independent, taking expectations of product of functions involving $\mathbf U_{\mathbf X}$ and $\mathbf U_{\mathbf H}$, $\mathbf U_{\mathbf X}$ and $\mathbf U_{\mathbf H}$, and, $\mathbf U_{\mathbf X}$ and $U_Y$ equals 0 (with the assumptions that $f_4(\mathbf U_{\mathbf H})=0$ etc.). 

The term $\underset{(\mathbf x,y)\sim P_i}{\mathbb E}\nabla_{\beta} g(\mathbf z;\beta_0)(y - g(\mathbf z;\beta_0))$ does not depend on $\mathbf U_{\mathbf X}$ nor the environment index such that, 
\begin{align}
    \underset{(\mathbf x,y)\sim P_i}{\mathbb E}\nabla_{\beta} g(\mathbf z;\beta_0)(y - g(\mathbf z;\beta_0)) - \underset{(\mathbf x,y)\sim P_j}{\mathbb E}\nabla_{\beta} g(\mathbf z;\beta_0)(y - g(\mathbf z;\beta_0)) = 0.
\end{align}

For the second result, consider again the Taylor expansion of $g(\mathbf z;\beta_0)$ around an estimate $\hat\beta$ sufficiently close to $\beta_0$, $g(\mathbf z;\beta_0) \approx g(\mathbf z;\hat\beta) + \nabla_{\beta} g(\mathbf z;\hat\beta)^T (\beta_0 - \hat\beta)$. Using this approximation, as in section \ref{sec_2} of the main body of this paper and our first order optimality condition we find,
\begin{align}
    \nabla_{\beta} g(\mathbf z;\hat\beta)\nabla_{\beta} g(\mathbf z;\hat\beta)^T(\hat\beta - \beta_0) + v = \nabla_{\beta} g(\mathbf z;\hat\beta) e,
\end{align}
where $v$ is a scaled disturbance term that includes the rest of the Taylor expansion of $g$ and is $\mathcal O((\hat\beta - \beta_0)^2)$ as $(\hat\beta - \beta_0)\rightarrow 0$. ; $e:= y - g(\mathbf z;\hat\beta)$ is the residual. Omitting $v$ and by taking the difference of this quantity estimated in two different environments $P_i$ and $P_j$ we get,
\begin{align*}
    \Big(\underset{(x,y)\sim P_i}{\mathbb E}\nabla_{\beta} g(\mathbf z;\hat\beta)\nabla_{\beta} g(\mathbf z;\hat\beta)^T & - \underset{(x,y)\sim P_j}{\mathbb E}\nabla_{\beta} g(\mathbf z;\hat\beta)\nabla_{\beta} g(\mathbf z;\hat\beta)^T\Big) (\hat\beta - \beta_0) \\ &=\left(\underset{(x,y)\sim P_i}{\mathbb E}\nabla_{\beta} g(\mathbf z;\hat\beta)e - \underset{(x,y)\sim P_j}{\mathbb E}\nabla_{\beta} g(\mathbf z;\hat\beta)e\right) \\
    &= 0.
\end{align*}

\subsection{Proof of Theorem 1}
\label{theorem_proof}

\textbf{Lemma 1.} \textit{Let $\beta_i \in (-W,W)$ be defined in a bounded open interval where $\beta_i$ is the $i$-th component of the parameter vector $\beta$. Then, the absolute deviation of the expected expected error in a given environment from the mean is bounded by the absolute deviation of its irreducible error from the mean and the difference in expected loss derivatives as follows,}
\begin{align*}
    \Big|\underset{(x,y)\sim P_e}{\mathbb E}  \mathcal L\left(f \circ \phi(x),y \right)&  -\underset{(x,y)\sim P_e, e\sim \mathcal E}{\mathbb E} \mathcal L\left(f \circ \phi(x),y \right) \Big| \leq \\ &\Big|\underset{y\sim P_e}{\mathbb E}  y^2 -\underset{y\sim P_e, e\sim \mathcal E}{\mathbb E} y^2 \Big| + 2W\cdot\Big|\hspace{0.1cm}\underset{\|\beta\|_{\infty}<W}{\sup}\hspace{0.1cm}\underset{(x,y)\sim P_e}{\mathbb E}  \frac{\partial}{\partial \beta_i}\mathcal L\left(f(\mathbf x),y \right) - \underset{(\mathbf x,y)\sim P_e, e\sim \mathcal E}{\mathbb E} \frac{\partial}{\partial \beta_i}\mathcal L\left(f(\mathbf x),y \right)\Big|.
\end{align*}

\textit{Proof.} By the Fundamental Theorem of Calculus, if $f$ is differentiable on a bounded open interval I (say $|x|\leq W$ for all $x\in I$) and its derivative is continuous on $I$ and bounded on $I$, then choosing some $a\in I$ we have,
\begin{align*}
    f(x) = f(a) + \int_{a}^x f'(t)dt
\end{align*}
By letting $f(\beta)= \underset{(x,y)\sim P_e}{\mathbb E}  \mathcal L\left(f \circ \phi(x),y \right)  -\underset{(x,y)\sim P_e, e\sim \mathcal E}{\mathbb E} \mathcal L\left(f \circ \phi(x),y \right)$, and $|\beta_i|\leq W$ it holds by the triangle inequality that,
\begin{align*}
    \Big|\underset{(x,y)\sim P_e}{\mathbb E}  &\mathcal L\left(f \circ \phi(x),y \right)  -\underset{(x,y)\sim P_e, e\sim \mathcal E}{\mathbb E} \mathcal L\left(f \circ \phi(x),y \right) \Big| \leq \\ &\Big|\underset{y\sim P_e}{\mathbb E}  y^2 -\underset{y\sim P_e, e\sim \mathcal E}{\mathbb E} y^2 \Big| + 2W\cdot\Big|\underset{\|\beta\|_{\infty}<W}{\sup}\hspace{0.1cm}\underset{(x,y)\sim P_e}{\mathbb E}  \frac{\partial}{\partial \beta_i}\mathcal L\left(f(\mathbf x),y \right) - \underset{(\mathbf x,y)\sim P_e, e\sim \mathcal E}{\mathbb E} \frac{\partial}{\partial \beta_i}\mathcal L\left(f(\mathbf x),y \right)\Big|,
\end{align*}
assuming that the partial derivative and the integral commute.

We restate the Theorem for convenience.

\textbf{Theorem 1} \textit{Let $\{P_e: e \in \mathcal E\}$ be a set of available environments indexed by $\mathcal E$. Further, assume the $\beta$ parameter space to be bounded $\|\beta\|_{\infty} \leq W$. Then, the following inequality holds,}
\begin{align*}
    \underset{\{\alpha_e\} \in \Delta_{\eta}}{\sup}\hspace{0.1cm} \sum_{e\in\mathcal E} \alpha_e \underset{(\mathbf x,y)\sim P_e}{\mathbb E} &\mathcal L\left(f(\mathbf x),y \right) \leq \underset{(\mathbf x,y)\sim P_e, e\sim \mathcal E}{\mathbb E}  \mathcal L\left(f(\mathbf x),y \right) \\
    &+ (1 + n\eta) \cdot 2W \cdot 
    \Big| \hspace{0.1cm} \underset{e\in \mathcal E, \|\beta\|_{\infty}<W}{\sup}\hspace{0.1cm}\underset{(\mathbf x,y)\sim P_e}{\mathbb E}  \nabla_{\beta}\mathcal L\left(f(\mathbf x),y \right) - \underset{(\mathbf x,y)\sim P_e, e\sim \mathcal E}{\mathbb E} \nabla_{\beta}\mathcal L\left(f(\mathbf x),y \right)\hspace{0.1cm} \Big |\\
     &+ (1 + n\eta) \cdot \Big|\hspace{.1cm}\underset{e\in \mathcal E}{\sup}\hspace{.1cm}\underset{y\sim P_e}{\mathbb E}  y^2 -\underset{y\sim P_e, e\sim \mathcal E}{\mathbb E} y^2 \Big|,
\end{align*}
\textit{where $n:= |\mathcal E|$ is the number of available environments and $e\sim\mathcal E$ loosely denotes sampling indices with equal probability from $\mathcal E$.}

\textit{Proof.} The following derivation shows the claim.
\begin{align*}
    \underset{\alpha_e \in \Delta_{\eta}}{\sup}\hspace{0.1cm} \sum_{e\in\mathcal E} &\alpha_e \underset{(x,y)\sim P_e}{\mathbb E} \mathcal L\left(f \circ \phi(x),y \right)  =  (1 + n\eta) \cdot \underset{e\in \mathcal E}{\sup}\hspace{0.1cm}\underset{(x,y)\sim P_e}{\mathbb E}  \mathcal L\left(f \circ \phi(x),y \right) \\
    &- \eta\sum_{e\sim \mathcal E} \mathbb E_{P_e}\mathcal L\left(f \circ \phi(x),y \right)\\
    &=  \underset{(x,y)\sim P_e, e\sim \mathcal E}{\mathbb E}  \mathcal L\left(f \circ \phi(x),y \right) + (1 + n\eta) \cdot \underset{e\in \mathcal E}{\sup}\hspace{0.1cm}\underset{(x,y)\sim P_e}{\mathbb E}  \mathcal L\left(f \circ \phi(x),y \right) \\ 
    &- (\eta + 1/n)\sum_{e\sim \mathcal E} \underset{(x,y)\sim P_e}{\mathbb E}\mathcal L\left(f \circ \phi(x),y \right)\\
    &= \underset{(x,y)\sim P_e, e\sim \mathcal E}{\mathbb E}  \mathcal L\left(f \circ \phi(x),y \right)\\
     &+ (1 + n\eta) \cdot  \Big ( \hspace{0.1cm} \underset{e\in \mathcal E}{\sup}\hspace{0.1cm}\underset{(x,y)\sim P_e}{\mathbb E}  \mathcal L\left(f \circ \phi(x),y \right)  -\underset{(x,y)\sim P_e, e\sim \mathcal E}{\mathbb E} \mathcal L\left(f \circ \phi(x),y \right) \Big )\\
     &\leq \underset{(x,y)\sim P_e, e\sim \mathcal E}{\mathbb E}  \mathcal L\left(f \circ \phi(x),y \right)\\
     &+ (1 + n\eta) \cdot 2W \cdot \Big\|\hspace{.1cm} \underset{e\in \mathcal E, \|\beta\|_{\infty}<W}{\sup}\hspace{0.1cm}\underset{(x,y)\sim P_e}{\mathbb E}  \nabla_\beta\mathcal L\left(f \circ \phi(x),y \right)  -\underset{(x,y)\sim P_e, e\sim \mathcal E}{\mathbb E} \nabla_\beta\mathcal L\left(f \circ \phi(x),y \right) \Big\|_{\infty} \\
     &+ (1 + n\eta) \cdot \Big|\hspace{.1cm}\underset{e\in \mathcal E}{\sup}\hspace{.1cm}\underset{y\sim P_e}{\mathbb E}  y^2 -\underset{y\sim P_e, e\sim \mathcal E}{\mathbb E} y^2 \Big|.
\end{align*}
The last inequality follows from Lemma 1. \QED

\newpage
\section{DomainBed comparisons}
\label{sec_domainbed}
% https://github.com/YugeTen/fish
% Gradient Matching for Domain Generalization

This section presents additional results on benchmark domain generalization tasks.

This section presents results on VLCS \cite{fang2013unbiased}, PACS \cite{li2017deeper} and Office-Home \cite{venkateswara2017deep} data sets using the DomainBed platform \cite{gulrajani2020search}. This allows us to make comparisons with a number of additional algorithms including Mixup \cite{yan2020improve}, MLDG \cite{li2018learning}, Coral \cite{sun2016deep} and MMD \cite{li2018domain}, all using a fixed and consistent hyperparameter selection procedure. We report average performance results on test data using a "training domain validation set" (i.e. a validation set is created by pooling together held-out subsets of data from each training domain) hyperparameter selection procedure as defined by \cite{gulrajani2020search}. We use default hyperparameter ranges, data augmentation and network architectures as defined in the DomainBed platform \cite{gulrajani2020search}.

The results are given in Table \ref{benchmarks}. The performance of DIRM is competitive on all datasets.

\begin{table}[H]
\fontsize{8}{9}\selectfont
\centering
\begin{tabular}{ |p{1.5cm}|C{0.9cm}|C{0.9cm}|C{0.9cm}|C{0.9cm}|C{0.9cm}|C{0.9cm}|C{0.9cm}|C{0.9cm}|C{0.9cm}| }
 \cline{2-10}
  \multicolumn{1}{c|}{} & ERM & Mixup & MLDG & Coral & MMD & DRO & IRM & REx & DIRM  \\
 \hline
 VLCS & 77.4  & 77.7    & 77.1  & 77.7  & 76.7  & 77.2 & 78.1  & 77.5  & 77.5  \\
 
 PACS & 85.6  & 84.4       & 84.8  & 86.0  & 85.1  & 84.1  & 84.3  & 84.0  & 84.6  \\
 
 Office-Home & 67.9 & 68.9 & 68.2 & 68.6 & 67.5 & 66.9  & 66.7  & 67.3  & 68.4 \\
 \hline
\end{tabular}
\caption{Training domain validation set accuracy in $\%$.}
\label{benchmarks}
\end{table}

\newpage
\section{Experimental details}
\label{experimental_details}
This section gives implementation details of all algorithms and a description of the medical data pre-processing.

\subsection{DIRM Implementation details}
\label{DIRM_implementation_details}

\begin{itemize}[leftmargin=*]
    \item \textbf{Regularization with $L_2$ norm.} The bound given in Theorem 1 quantifies the discrepancy between function derivatives using the $L_2$ norm, defined as an integral over possible parameter values $\beta$. For neural networks, computation of the $L_2$ norm is largely intractable and specifically, for networks of depth greater or equal to $4$, it is an NP-hard problem (see Proposition 1 in \cite{triki2017function}). Some approximation is thus unavoidable. 

    One option is to recognise the $L_2$ norm as an expectation over functional evaluations, $||f||_{L_2} = \mathbb E_{x\sim\mathcal U(\Theta)}\left [||f(x)||_2^2\right]^{1/2}$ for a continuous function $f$ taking values $x$ sampled uniformly from its domain $\Theta$. Empirical means are tractable yet they induce a much higher computational burden as these must be computed in every step of the optimization since $\phi$ is changing. Our approach is to take this approximation to its limit, making a single function evaluation at each step of the optimization using the current estimate $\beta$, as written in Algorithm 1. 

    This approximation loosens the connection between the bound given in Theorem 1 and the proposed algorithm. It remains justified however from a conceptual and empirical perspective. Conceptually, the objective of controlling an $L_2$ type of norm is to encourage the regularizer function towards 0, and thus the values of the regularizer (which we choose to do explicitly). Empirically, we make performance comparisons with the alternative of explicitly computing empirical means over a grid of parameter evaluations. we implemented empirical means using all combinations of parameter values chosen from a grid of $5$ parameter values around the current estimate $\beta$, $\{0.25\beta, 0.5\beta, \beta, 2\beta, 4\beta\}$. Table \ref{approximation} shows similar performance across the real data experiments considered in the main body of this paper. Our conclusion is that a single evaluation is in practice enough to monitor invariance of representations to environment-specific loss derivatives.  
    
    \item \textbf{Optimization.} Pseudocode for DIRM is given in Algorithm 1.
    
    \begin{center}
\begin{minipage}{.85\linewidth}
\begin{algorithm}[H]
%\fontsize{9}{9}\selectfont
   \caption{DIRM}
   \label{alg:DIRM}
\begin{algorithmic}
   \STATE {\bfseries Input:} datasets $\mathcal D_1,\dots,\mathcal D_E$ in $E$ different environments, parameter $\lambda$, batch size $K$
   \STATE {\bfseries Initialize:} neural network model parameters $\phi,\beta$ 
   \WHILE{convergence criteria not satisfied}
   \FOR{$e = 1,\dots,E$}
   \STATE Estimate loss $\mathcal L_e(\phi,\beta)$ empirically using a batch of $K$ examples from $\mathcal D_e$.
   \STATE Estimate derivatives $\nabla_{\beta}\mathcal L_e(\phi,\beta)$ empirically using a batch of $K$ examples from $\mathcal D_e$.
   \ENDFOR
   \STATE Update $\beta$ by stochastic gradient descent with,
   \[\nabla_{\beta}\left( \frac{1}{E}\sum_{e=1}^E \mathcal L_e(\phi,\beta)\right)\]
   \STATE Update $\phi$ by stochastic gradient descent with,
   \begin{align*}
       \nabla_{\phi}\Big( \frac{1}{E}\sum_{e=1}^E \mathcal L_e(\phi,\beta) + \lambda\cdot \text{Var}(||\nabla_{\beta}\mathcal L_1(\phi,\beta)||_2^2, \dots, ||\nabla_{\beta}\mathcal L_E(\phi,\beta)||_2^2)\Big)
   \end{align*}
   \ENDWHILE
\end{algorithmic}
\end{algorithm}
\end{minipage}
\end{center}

    \item \textbf{Initialization and hyperparameters.} DIRM is sensitive to initialization and to the choice of hyperparameters -- specifically its optimization schedule. In our experiments, we found best performance by increasing the relative weight of the penalty term $\lambda$ after a fixed number of iterations (and similar implementations are used for IRM and REx that suffer from similar challenges). This we believe could be a significant limitation for its use in practice since this choice must be made a priori. We investigated the sensitivity of DIRM to this optimization schedule in Table \ref{scheduling} that shows test accuracy as a function of the iteration at which penalty term weight $\lambda$ is increased. 
    
    Choosing this number accurately is important for generalization performance. If $\lambda$ is increased too early, different initialization values (and the complex loss landscape) lead to different solutions with unreliable performance and a large variance. This happens for all methods. An initial number of iterations minimizing loss in-sample improves estimates for all methods which then converge to solutions that exhibit lower variance.
\end{itemize}

\begin{table}[H]
\fontsize{9.5}{10.5}\selectfont
\centering
\begin{tabular}{ |p{3cm}|C{3cm}|C{3cm}|C{2.5cm}|  }
 \cline{2-4}
  \multicolumn{1}{c|}{} & \textbf{Pneumonia Prediction} & \textbf{Parkinson Prediction} & \textbf{Survival Prediction} \\
 \hline
 DIRM (mean approximation of $L_2$ norm) & 63.5 ($\pm$ 3) & 73.0 ($\pm$ 1.5) & 78.0 ($\pm$ .9) \\
 DIRM & 63.7 ($\pm$ 3) & 72.8 ($\pm$ 2) & 77.9 ($\pm$ 1)\\
 \hline
\end{tabular}
\caption{Test set performance (accuracy in $\%$) on real datasets for two different regularizer approximations to the $L_2$ norm.}
\label{approximation}
\end{table}

\begin{table}[H]
\fontsize{9.5}{10.5}\selectfont
\centering
\begin{tabular}{ |p{1cm}|C{1.6cm}|C{1.6cm}|C{1.6cm}|C{1.6cm}|C{1.6cm}|C{1.6cm}|  }
 \cline{2-7}
  \multicolumn{1}{c|}{} & \textbf{2 epochs} & \textbf{4 epochs} & \textbf{6 epochs} & \textbf{8 epochs} & \textbf{10 epochs} & \textbf{12 epochs}  \\
 \hline
 IRM & 56.4 ($\pm$ 6) & 58.1 ($\pm$ 3) & 59.2 ($\pm$ 3) & 59.1 ($\pm$ 2) & 58.1 ($\pm$ 3) & 57.8 ($\pm$ 1) \\
 REx & 55.3 ($\pm$ 8) & 57.9 ($\pm$ 4) & 60.6 ($\pm$ 3) & 60.5 ($\pm$ 2) & 57.7 ($\pm$ 3) & 54.7 ($\pm$ 2) \\
 DIRM & 54.1 ($\pm$ 7) & 61.7 ($\pm$ 4) & 63.8 ($\pm$ 3) & 63.5 ($\pm$ 3) & 62.6 ($\pm$ 2) & 58.2 ($\pm$ 1) \\
 \hline
\end{tabular}
\caption{Test set performance (accuracy in $\%$) on X-ray data as a function of the number of epochs used to increase penalty $\lambda$.}
\label{scheduling}
\end{table}

\subsection{Baseline implementation details}

All algorithms are implemented with the same architecture, activation functions, hyperparameter optimization procedures and will depend to some extent on the modality of the data. These details are given in the section on data description. 

We use our own implementation of ERM and DRO, and use the code at \texttt{https://github.com/fungtion/DANN} for DANN, the code at \texttt{https://github.com/facebookresearch/InvariantRiskMinimization} for IRM, and the code at \texttt{https://github.com/capybaralet/REx\_code\_release} for REx.

\subsection{X-ray Data}
\label{x_ray_data_details}

\textbf{Source and pre-processing.} The X-ray dataset from the Guangzhou Women and Children’s Medical Center can be found at \texttt{https://www.kaggle.com/paultimothymooney/chest-xray-pneumonia} and the X-ray dataset from the National Institutes of Health can be found at \texttt{https://www.kaggle.com/nih-chest-xrays/data}. We will release all pre-processing details upon acceptance of the paper. We describe them briefly next.

We create training environments with different proportions of $X$-rays from our two hospital sources to induce a correlation between the hospital (and its specific data collection procedure) and the pneumonia label. The objective is to encourage learning principles to exploit a spurious correlation, data collection mechanisms should not be related to the probability of being diagnosed with pneumonia. The reason for creating two training data sets with slightly different spurious correlation patterns is to nevertheless leave a statistical footprint in the distributions to disentangle stable (likely causal) and unstable (likely spurious). In each of the training and testing datasets we ensured positive and negative labels remained balanced. 

The training datasets contained 2002 samples each and the testing dataset contained 1144 samples.

\textbf{Architecture of networks for this experiment.} All learning paradigms trained a convolutional neural network, 2 layers deep, with each layer consisting of a convolution (kernel size of 3 and a stride of 1). The number of input channels was 64, doubled for each subsequent layer, dropout was applied after each layer and the elu activation function was used. We optimize the binary cross-entropy loss using Adam without further regularization on parameters and use Xavier initialization. 

Learning rates and other hyperparameters (e.g. $\lambda$ value and epoch schedule for raising the regularization term from 0 to $\lambda$ for DIRM, IRM and REx) are chosen on a held-out validation set mimicking the procedure of DomainBed \cite{gulrajani2020search}: a validation set is created by pooling together held-out subsets of data from each training domain. Experiments are run for a maximum of 20 epochs with early stopping based on validation performance. All results are averaged over 10 trials with different random splits of the data, and the reported uncertainty intervals are standard deviations of these 10 performance results.

\subsection{Parkinson's Disease Speech Data}
\label{parkinson_data_details}

\textbf{Source and pre-processing.} The Parkinson's data can be found at \texttt{https://archive.ics.uci.edu/ml/datasets/parkinsons}. The data includes a total of 26 features recorded on each sample of speech and set training and testing splits which we use in our experiments. For each patient 26 different voice samples including sustained vowels, numbers, words and short sentences where recorded, which we considered to be different but related data sources. We created three training environments by concatenating features from three number recordings, concatenating features from three word recording and concatenating features from three sentences; for a total of 120 samples in each of the three training environment. The available testing split contained 168 recordings of vowels, which we expect to differ from training environments because these are different patients and do not contain numbers or words. Positive and negative samples were balanced in both training and testing environments. 

\textbf{Architecture of netowrks for this experiment.} On this data, for all learning paradigms we train a feed-forward neural network with two hidden layers of size 64, with elu activations and dropout $(p=0.5)$ after each layer. As in the image experiments, we optimize the binary cross-entropy loss using Adam, $L_2$ regularization on parameters and use Xavier initialization. 

Learning rates and other hyperparameters (e.g. $\lambda$ value and epoch schedule for raising the regularization term from 0 to $\lambda$ for DIRM, IRM and REx) are chosen on a held-out validation set mimicking the procedure of DomainBed \cite{gulrajani2020search}: a validation set is created by pooling together held-out subsets of data from each training domain. Experiments are run for a maximum of 1000 epochs with early stopping based on the validation performance. All results are averaged over 10 trials with different random seeds of our algorithm. This is to give a sense of algorithm stability rather than performance stability. 

\subsection{MAGGIC Electronic Health Records}
\label{maggic_data_details}

\textbf{Source and pre-processing.} MAGGIC stands for Meta-Analysis Global Group in Chronic Heart Failure. The MAGGIC meta-analysis includes individual data on $39,372$ patients with heart failure (both reduced and preserved left-ventricular ejection fraction), from 30 cohort studies, six of which were clinical trials. It is privately held. $40.2\%$ of patients died during a median follow-up of 2.5 years. For our purposes, we removed patients that were censored or lost to follow-up to ensure well-defined outcomes after 3 years after being discharged from their respective hospitals. A total of 33 variables describe each patient including demographic variables: age, gender, race, etc; biomarkers: blood pressure, haemoglobin levels, smoking status, ejection fraction, etc; and details of their medical history: diabetes, stroke, angina, etc.

On all patients follow-up over 3 years, we estimated feature influence of survival status after three years. A number of variables were significantly associated with survival out of which we chose Age, also found correlated with BMI and a number of medical history features, as a confounder for the effect of these variables on survival. We used three criteria to select studies: having more than 500 patients enrolled and balanced death rates (circa $50\%$). 5 studies fitted these constraints: 'DIAMO', 'ECHOS', 'HOLA', 'Richa', 'Tribo'. Each was chosen in turn as a target environment with models trained on the other 4 training environments.

\textbf{Architecture of networks for this experiment.} The same architecture and hyperparameters as in Parkinson's disease speech data experiments was used for MAGGIC data except that we increase the maximum training epochs to 5000. Learning rates and other hyperparameters (e.g. $\lambda$ value and epoch schedule for raising the regularization term from 0 to $\lambda$ for DIRM, IRM and REx) are chosen on a held-out validation set mimicking the procedure of DomainBed \cite{gulrajani2020search}: a validation set is created by pooling together held-out subsets of data from each training domain. Experiments are run for a maximum of 1000 epochs with early stopping based on the validation performance. All results are averaged over 10 trials with different random seeds of our algorithm.

\textbf{Feature reproducibility experiments}. A natural objective for the consistency of health care and such that we may reproduce the experiments and their results in different scenarios is to find relevant features that are not specific to an individual medical study, but can also be found (replicated) on other studies with different patients. Heterogeneous patients and studies, along with different national guidelines and standards of care make this challenging. 

In our experiments we made comparisons of reproducibility in parameter estimates for models trained using ERM and DIRM. We chose networks with a single layer with logistic activation and focused on the estimation of parameter to understand the variability in training among different data sources. Naturally, feature importance measured by parameter magnitudes makes sense only after normalization of the covariates to the same (empirical) variance (equal to 1) in each study separately. After this pre-processing step, for both ERM and the proposed approach we trained separate networks on 100 random pairs of studies (each pair concatenated for ERM) and returned the top 10 significant features (by the magnitude of parameters). Over all sets of significant parameters we then identified how many intersected across a fixed number of the 100 runs. 

\end{document}